\documentclass{article}

\usepackage{PRIMEarxiv}
\usepackage{xcolor}

\usepackage[utf8]{inputenc} 
\usepackage[T1]{fontenc}    
\usepackage{hyperref}       
\usepackage{url}            
\usepackage{booktabs}       
\usepackage{amsfonts}       
\usepackage{nicefrac}       
\usepackage{microtype}      
\usepackage{lipsum}
\usepackage{fancyhdr}       
\usepackage{graphicx}       
\graphicspath{{images/}}    
\usepackage{listings}       
\usepackage{spverbatim}
\usepackage{float}          

\usepackage[numbers]{natbib}

\usepackage{tabulary}
\usepackage{array,booktabs,ragged2e}
\newcolumntype{R}[1]{>{\RaggedLeft\arraybackslash}p{#1}}
\newcolumntype{C}[1]{>{\Centering\arraybackslash}p{#1}}

\title{Arctic Long Sequence Training: Scalable And Efficient Training For Multi-Million Token Sequences}

\author{
  Stas Bekman, Samyam Rajbhandari, Michael Wyatt, Jeff Rasley, \\
  Tunji Ruwase, Zhewei Yao, Aurick Qiao and Yuxiong He\\
  Snowflake AI Research \\
  \texttt{\{stas.bekman, samyam.rajbhandari, michael.wyatt, jeff.rasley\}@snowflake.com} \\
  \texttt{\{tunji.ruwase, zhewei.yao, aurick.qiao, yuxiong.he\}@snowflake.com} \\
}

\begin{document}
\maketitle
\begin{abstract}
\bfseries{

Long sequences are critical for applications like RAG, long document summarization, multi-modality, etc., and modern LLMs, like Llama 4 Scout, support max sequence length of up to 10 million tokens. However, outside of enterprise labs, long sequence training is challenging for the AI community with limited system support in the open-source space. 

Out-of-box, even on a modern NVIDIA H100 80GB GPU cluster, training Llama 8B model with sequence over 32K runs out of memory on a basic Hugging Face (HF) model due to two reasons: i) LLM training workloads are not optimized to fully leverage a single GPU memory, ii) existing solutions for leveraging multiple GPU memory are not easily available to HF models, making long sequence training inaccessible.

We address this with Arctic Long Sequence Training (ALST). It offers a combination of attention-agnostic single GPU and multi-GPU memory optimizations, that enables it to support out-of-box training of multi-million sequence length for a wide variety of HF models.

ALST supports training Meta's Llama 8B model with 500K sequence length on a single H100 GPU, 3.7M on a single 8xH100 GPU node, and over 15M on a 4 node cluster, an increase of over 400x compared to the 32K baseline for the latter. ALST is fully compatible with HF models and open-sourced via \href{https://www.deepspeed.ai/tutorials/ulysses-alst-sequence-pallellism/}{DeepSpeed} and \href{https://github.com/snowflakedb/ArcticTraining/blob/main/projects/sequence-parallelism/README.md}{Arctic Training}.
}
\end{abstract}


\section{Introduction}

Long sequence capability offers the ultimate unlock for a wide range of
AI applications from RAG, multi-turn conversation, long document
summarization, multi-modality support, and many more. This is evident
from the continuous increase in the max sequence length supported by
popular Open Source LLMs, like Meta's Llama 3.x and Alibaba's Qwen 2.5 32B,
which support 128K-token sequences, NVIDIA's Llama-3.1-8B-UltraLong-4M-Instruct finetuned to support a 4M
sequence length, and the more recent Meta's Llama-4 Maverick and Llama 4-Scout models supporting
1M and and a whopping 10M sequence length, respectively.

While these models are capable of handling incredibly long sequence lengths, for
several reasons, fine-tuning these models at these sequence lengths to
enhance task-specific capabilities is out of reach for most data
scientists who do not have access to sophisticated enterprise training
systems and rely on open source solutions:

\begin{itemize}
\item First, standard LLM training workflows are not optimized for memory efficiency, which restricts the maximum sequence length per GPU and makes them suboptimal for long-sequence training.
\item Second, training with multi-million sequence lengths requires more
  memory than available in any commercially available GPU devices, and
  while there are solutions that allow for leveraging aggregate GPU
  memory across multiple devices, they are limited. For example, Ring-based sequence parallelism ~\cite{liu2023ringattentionblockwisetransformers} does not support arbitrary attention patterns natively. While Ulysses-based sequence parallelism~\cite{jacobs2023deepspeed} does not
  have this restriction, it is not supported in popular frameworks like
  Hugging Face, limiting its accessibility.
\item Third, PyTorch itself suffers from a multitude of memory bottlenecks
  that limit the available memory available to support long sequences.
\end{itemize}

In the Open Source release of Arctic Long Sequence Training (ALST) we address the above challenges with three targeted solutions:

\begin{itemize}
\item Ulysses Sequence Parallelism Compatible with Hugging Face Transformers: Adapted from the original Megatron-DeepSpeed Ulysses \cite{jacobs2023deepspeed} and extended to support modern Attention mechanisms, this technique enables the use of aggregate GPU memory across multiple devices.

\item Sequence Tiling for Memory Efficiency: A new computation tiling pattern for LLM training that reduces the memory required for memory intensive operators like logits, loss, and MLP computation from O(N) to O(1), where N is the sequence length.

\item PyTorch Memory Optimizations: Through comprehensive memory profiling of long-sequence training workloads, we identified and applied a series of PyTorch-specific optimizations to eliminate the unnecessary memory overheads.
\end{itemize}

By leveraging these three components and making them compatible with Hugging Face Transformers, ALST makes long-sequence training accessible to the broader AI community:

\begin{itemize}
\item 500K-long sequence training on a single H100 GPU, 16 times longer than baseline\footnote{The best prior to this work setup we found is explained in section \ref{feature-ablations}.\label{best-baseline-setup}}, democratizing long-sequence training on resource constrained setups.
\item 3.7M-long sequence training on a single H100 node, with a 116 times improvement relative to baseline\footref{best-baseline-setup}.
\item 15M-long sequence training on four H100 GPU node cluster, with a 469 times improvement relative to baseline\footref{best-baseline-setup}.
\item The technology is agnostic to the attention mechanism, allowing for out-of-box support for different sparsity patterns like block sparse, MoBA, etc.
\end{itemize}

\begin{figure}[H]
  \centering
  \includegraphics[width=0.7\textwidth]{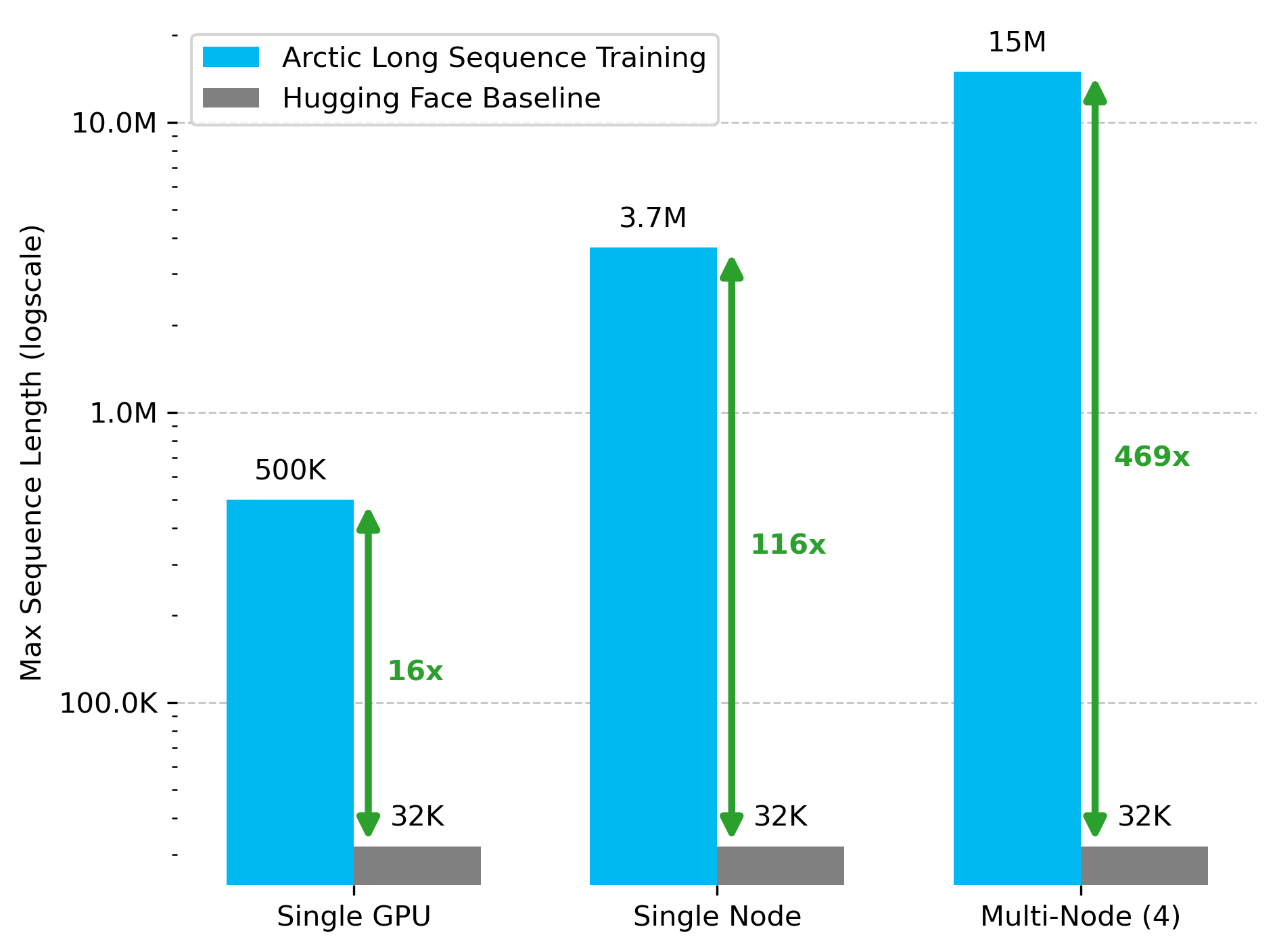}
  \caption{A dramatic improvement in sequence length with ALST enabled on 1, 8 and 32 H100 GPUs with Llama-8B. The baseline is Hugging Face with DeepSpeed ZeRO Stage 3 and optimizer states offload to CPU}
  \label{fig:fig1}
\end{figure}

Figure \ref{fig:fig1} provides a quick preview of the ALST accomplishments, which we will discuss in detail in later sections. \footnote{A log scale had to be used in order to visualize the baseline, since otherwise for the 469x improvement it won't even show up.}

Besides Ulysses SP~\cite{jacobs2023deepspeed} there are other approaches to sequence parallelism (also referred to as context parallelism). One approach was introduced by Megatron-LM~\cite{korthikanti2022reducingactivationrecomputationlarge} which extends Tensor Parallelism (TP) and which cannot operate without TP. And the more popular one, which can be used alone, is Ring Attention with many variants~\cite{liu2023ringattentionblockwisetransformers}~\cite{li2022sequenceparallelismlongsequence}~\cite{li2024distflashattndistributedmemoryefficientattention}~\cite{brandon2023stripedattentionfasterring}, which resembles a distributed version of Flash Attention 2 \cite{dao2023flashattention2fasterattentionbetter}. Some techniques combine the Ring Attention and Ulysses SP approaches~\cite{fang2024uspunifiedsequenceparallelism}~\cite{gu2024loongtrainefficienttraininglongsequence}. The main obstacle with these approaches is that they require modeling code modifications whereas Ulysses SP is attention and model-agnostic.

In the rest of the paper, we will take a deeper dive into the memory
challenges of long-sequence training, followed by a discussion of memory
optimizations targeting these challenges. For our readers interested in
deeper discussions, we also share our implementation details as well as
the nuances of integration into Hugging Face Transformers. Finally, we present our evaluation
results, as well as share how you can get started with ALST. For data
scientists looking to experiment with long-sequence training, we will
also share limitations and useful notes at the end of the paper.

\hypertarget{why-is-training-with-long-sequences-challenging}{%
\section{\texorpdfstring{Why is training with long sequences challenging?
}{Why is Training with Long Sequences Challenging?}}\label{why-is-training-with-long-sequences-challenging}}

Training models on long sequence lengths is a difficult task because
models are large and the accelerator memory is typically insufficient to
hold the large activations, especially when weights, optimizer states and gradients already consume a lot of the available GPU memory.

We used a combination of the
\href{https://pytorch.org/docs/main/torch_cuda_memory.html}{\underline{PyTorch memory profiler}} and a helper
\href{https://github.com/deepspeedai/DeepSpeed/blob/8f93f8b9b003ffec4a7c531a7997b0cb945e91bb/deepspeed/runtime/utils.py\#L771-L792}{\underline{see\_memory\_usage}} utility, that dumped memory usage stats at various places in the code, to identify
where memory was allocated in a non-efficient way or not released soon enough.

\hypertarget{model-training-memory-map}{%
\subsection{Model Training Memory Map}\label{model-training-memory-map}}

Here is a detailed breakdown of what the GPU memory is used for:

\begin{enumerate}
\def\labelenumi{\arabic{enumi}.}
\item \textbf{Weights + Optimizer states + Gradients}: In a typical BF16 mixed
  precision training about 18 bytes are needed per model parameter just
  to hold the model weights (2), optimizer states (8+4) and gradients
  (4). For example,
  \href{https://huggingface.co/meta-llama/Llama-3.1-8B-Instruct}{\underline{Llama-3.1-8B-Instruct}}
  contains 8 billion parameters and thus requires 16GiB for BF16 weights,
  64GiB for Adam optimizer states, 32GiB for FP32 weights used for
  optimizer stability, and finally 32GiB for FP32 gradients. Therefore
  in total each GPU already requires 144GiB of memory to train this 8B-parameter
  model before all the other overheads. \footnote{There are many
  \href{https://github.com/stas00/ml-engineering/blob/master/training/performance/README.md\#anatomy-of-models-memory-usage}{\underline{mixed
  precision training recipes}} but we find the one we use to be the most
  practical.}
\item \textbf{Activations}: Then there is memory required to calculate and
  hold activations - which is all the intermediate tensors that the
  model uses. This includes a variety of temporary tensors as well. The
  tricky part about activation memory is getting tensors that are no
  longer needed released as soon as possible. For example, activation
  checkpointing is often used to help reduce the required active memory
  by recalculating the intermediate forward activations during the
  backward call.
\item \textbf{Runtime overheads}: We observe that the remaining GPU memory
  is consumed by several key sources. Lower-level libraries such as CUDA
  and NCCL each reserve a significant amount of memory---CUDA typically
  uses around 1 GiB per GPU, while NCCL can consume multiple gigabytes
  for internal buffers and data structures, which are more difficult to
  track\footnote{PyTorch memory profiler doesn't track NCCL internal memory allocations.}. Additionally, different versions of PyTorch can exhibit varying
  memory usage patterns due to leaks or inefficiencies. Finally, when
  operating near the limits of available GPU memory, memory
  fragmentation becomes a concern, reducing the availability of large
  contiguous memory blocks.
\end{enumerate}

To better understand memory requirements for different models, GPU
counts, and total sequence length, we suggest 
\href{https://deepspeed.readthedocs.io/en/latest/memory.html}{\underline{the
API to estimate memory usage in DeepSpeed ZeRO}}.\footnote{See also:
\href{https://model-training-memory-calculator.streamlit.app/}{\underline{Interactive training memory
calculator}}.}.

\hypertarget{activation-memory-is-the-primary-bottleneck}{%
\subsection{\texorpdfstring{Activation Memory is the Primary
Bottleneck}{ Activation Memory is the Primary Bottleneck}}\label{activation-memory-is-the-primary-bottleneck}}

\begin{figure}[H]
  \centering
  \includegraphics[width=0.75\textwidth]{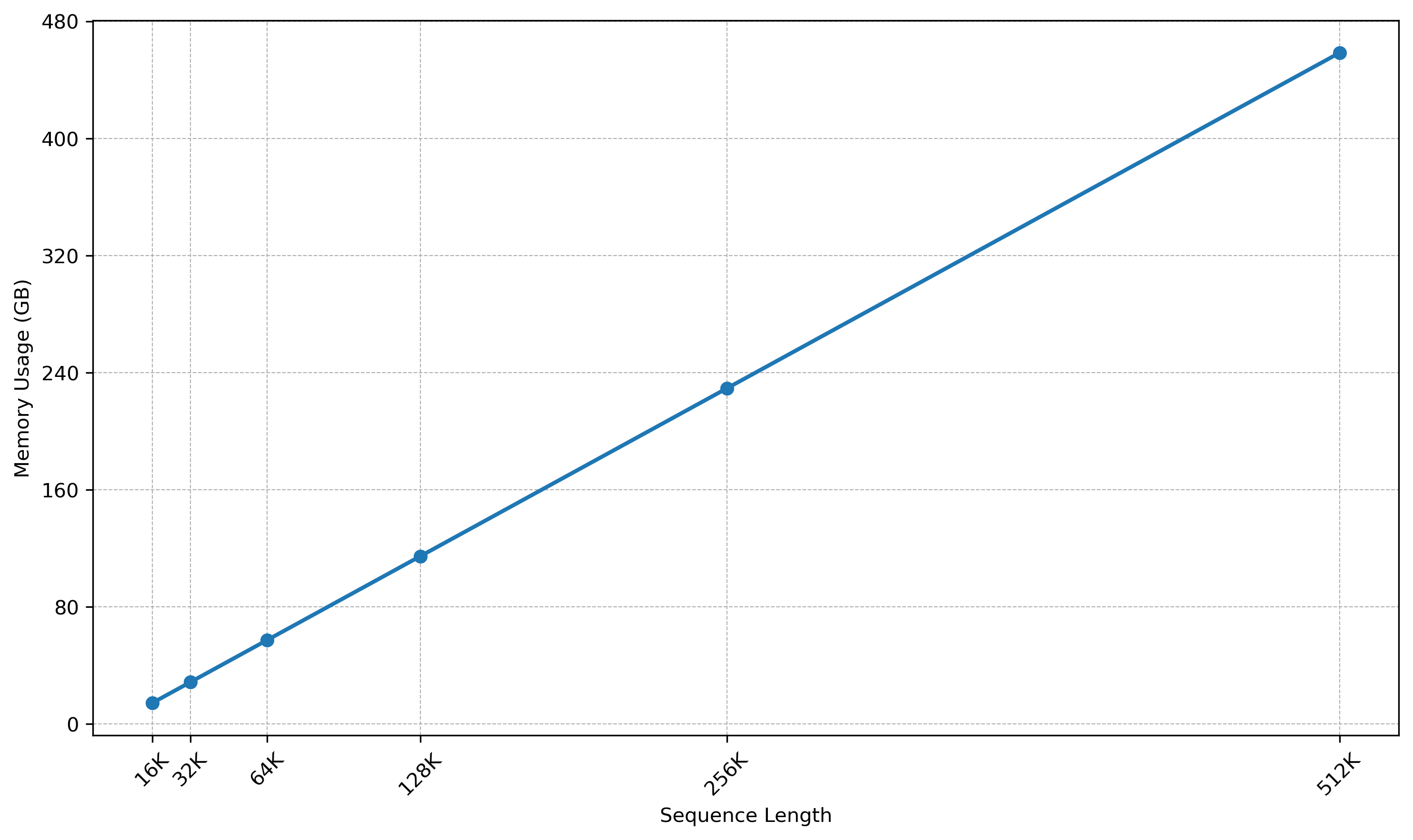}
  \caption{Estimated memory usage for Llama-8B activation memory with different sequence lengths.}
  \label{fig:fig3}
\end{figure}

Now that you understand what GPU memory is used for it should be easy to
understand that in order to go from a short sequence length to a long
one it's mainly more of the activation memory that we need to fit. All
other memory allocations remain the same regardless of the sequence
length. Figure \ref{fig:fig3}\footnote{Compiled with the help of \href{https://model-training-memory-calculator.streamlit.app/}{\underline{Interactive training memory
calculator}}.} shows how activation memory for Llama-8B increases linearly with sequence length.\footnote{Memory here is activation checkpoints memory + activation and logits working memory.}.

\hypertarget{memory-optimizations}{%
\section{\texorpdfstring{Memory Optimizations
}{Memory Optimizations }}\label{memory-optimizations}}

Next we will discuss three memory optimization groups that were instrumental at enabling training at very long sequence lengths:

\begin{enumerate}
\item Sequence tiling for reducing activation memory footprint
\item Ulysses Sequence Parallelism for Cross-GPU Activation Memory Sharing
\item Activation offloading and other PyTorch optimizations
\end{enumerate}

\hypertarget{sequence-tiling-for-reducing-activation-memory-footprint}{%
\subsection{Sequence Tiling for Reducing Activation Memory Footprint}\label{sequence-tiling-for-reducing-activation-memory-footprint}}

GPU memory requirements for training on long sequences grow rapidly
with sequence length increase. As part of our activation memory calculations~\ref{activation-memory-is-the-primary-bottleneck}, we
estimated the activation and logits memory needed for various models and
sequence lengths. Without our optimizations, the per-GPU memory usage
quickly becomes unsupportable~\footnote{i.e. hitting the Out-Of-Memory event.}---as shown in Figure \ref{fig:fig3}---and this
doesn't even include model parameters or optimizer states discussed
earlier.

To address this memory explosion, we leverage \textit{Sequence Tiling}, a technique that reduces peak memory usage by tiling forward and backward computations along the sequence dimension. Instead of processing the entire sequence in a single pass --- which requires storing large intermediate tensors --- Sequence Tiling breaks the computation into smaller tiles. Intermediate values are materialized only for each tile, significantly lowering the memory footprint.

This approach is applicable to operations that have no cross-sequence dependencies, such as linear layers, token embeddings, and per-token loss computations. For example, instead of computing logits or applying MLP layers across the entire sequence at once, we apply them sequentially to smaller segments, storing only the necessary intermediates at each step.

Let's start by examining how effective Sequence Tiling is at reducing
memory overhead during loss calculations. Using the example of
\href{https://huggingface.co/meta-llama/Llama-3.1-8B-Instruct/blob/main/config.json}{\underline{Llama-3.1-8B-Instruct}}
with a sequence length of 16K, the model's vocabulary size of 128,256
results in a single copy of the logits in FP32 consuming approximately ~8GiB of memory per GPU (calculated as \(4\times16\_000\times128\_256/2^{30}=7.65GiB)\). Rather
than materializing all 8GiB of logits at once for both the forward and
backward passes, we shard the logits into smaller chunks and compute the
forward and backward passes on each shard independently. This
significantly reduces peak memory usage. For instance, using a 1GiB
shard size divides the computation into about 8 chunks and can save over
14GiB of memory in practice (because the loss computation uses 2 times of 8GiB).

\begin{figure}[H]
  \centering
  \includegraphics[width=1\textwidth]{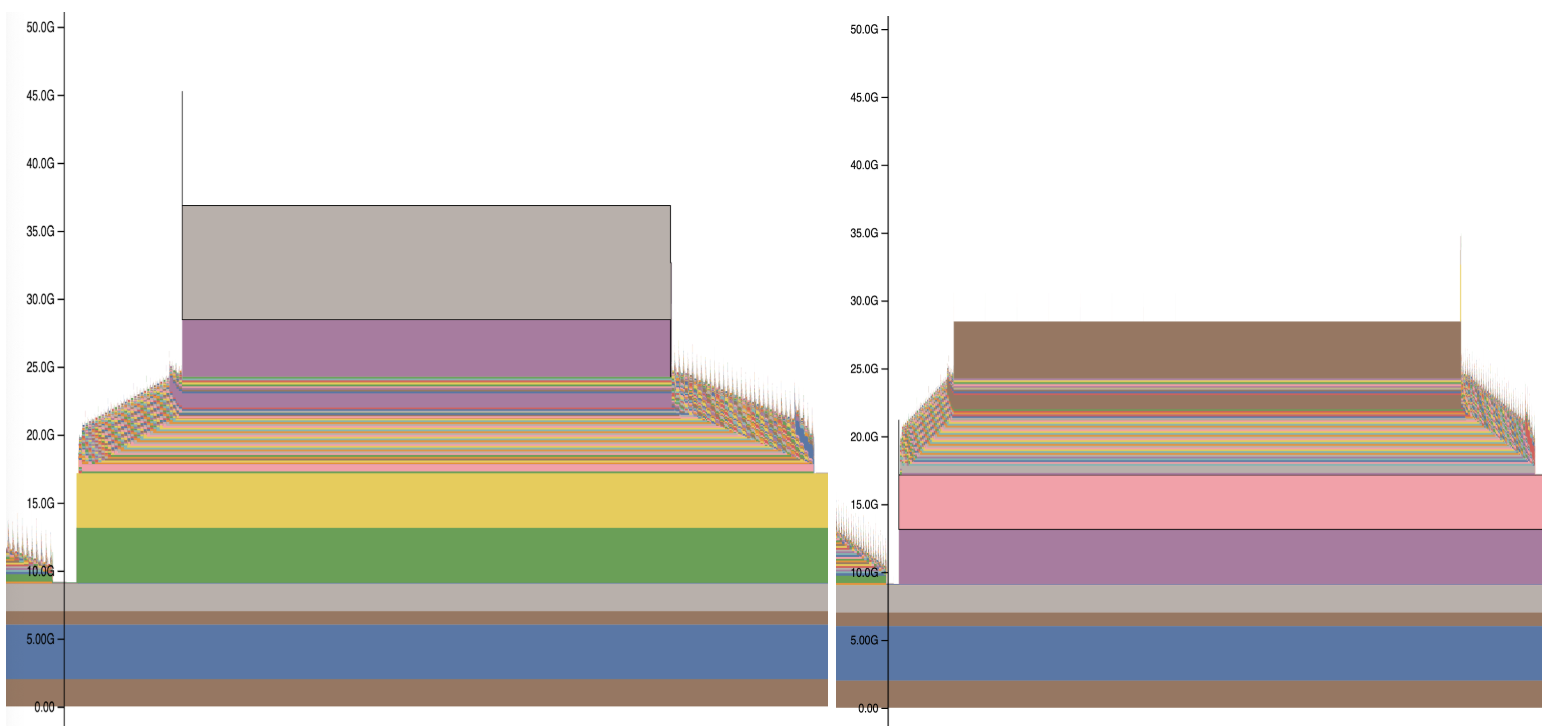}
  \caption{PyTorch memory usage plots before (left) and after (right) using Sequence Tiling to reduce loss calculation memory usage}
  \label{fig:fig4}
\end{figure}

Up to this point we have discussed memory reductions from a theoretical
perspective, we can also showcase these memory reductions empirically
via the help of the PyTorch memory profiler. In Figure \ref{fig:fig4} we show
two plots: (left) without Sequence Tiling the loss calculation we see
peak memory usage is around 50 GiB compared to (right) after updating the
loss calculation to use Sequence Tiling we see the peak memory usage
drops to 36 GiB which results in a 28\% memory reduction.
\footnote{There is a very thin spike that goes to 49.5GiB left and 36GiB (right) but it can't be seen in the small snapshot of the memory profiler plot.}
\footnote{The PyTorch memory profiler assigns colors at random, that's why most
  block colors don't match between the left and the right.}

This is not the first time a tiled compute is used. Here are some examples of recent use relevant to our discussion:

\begin{itemize}
\item For some years DeepSpeed's TiledLinear has been enabling much bigger
  compute loads that otherwise would have not fit into the GPU memory.
\item \href{https://github.com/linkedin/Liger-Kernel}{\underline{Liger-Kernel}} \cite{hsu2025ligerkernelefficienttriton}
  implemented a fused cross entropy loss, without manifesting the whole logits
  tensor first, thus enabling bigger batches and sequence lengths, but
  only for select most popular model architectures.
\end{itemize}

Now we introduce a generic TiledCompute autograd function that in theory
should be able to make any function that performs large matrices multiplications use much less GPU memory. We implemented a fused cross-entropy using it, but Liger-Kernel's version
of the same is a bit faster since it uses a Triton kernel. Liger-Kernel supports a limited number of popular Hugging Face Transformers architectures, but our solution in theory
should work with any architecture, so we recommend using Liger-Kernel's
fused cross entropy for the architectures it supports and when it
doesn't, then you have the option of using our implementation.

\subsubsection{TiledMLP}

But why stop at tiling the logits+loss computation, we can also tile the MLP compute\footnote{The attention block can't be tiled because it needs the whole sequence to attend to.}. We created a simplified version of TiledCompute and called it TiledMLP\footnote{TiledMLP is almost the same as TiledCompute but its code is easier to understand as it's not generalized.} to also perform a much more memory-efficient MLP computation, which allowed a huge increase in the sequence length. 

\begin{figure}[H]
  \centering
  \includegraphics[width=0.9\textwidth]{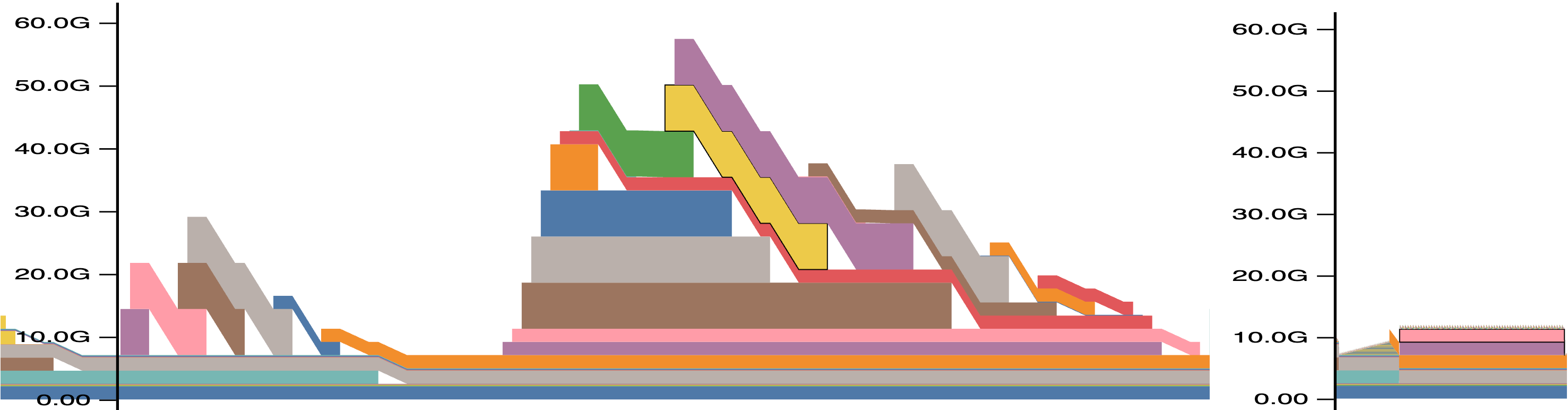}
  \caption{1x forward-backward cycle on a single Llama-8B LlamaMLP layer (Left) without tiling (Right) with tiling.}
  \label{fig:mem-profiler-tiled-mlp-llama-8b-1layer}
\end{figure}

If we extract a single LlamaMLP layer from Llama-8B and run a bf16 hidden\_states tensor of shape [1, 256\_000, 4096] through its forward-backward, without and with sequence dimension tiling, we get about 10x memory saved as can be seen from Figure \ref{fig:mem-profiler-tiled-mlp-llama-8b-1layer}\footnote{10-60GiB vs 7-12GiB memory used for forward-backward compute, the rest are static invariant structures.}. The number of shards was auto-deduced via \(ceil(seqlen=256\_000 / hidden\_size=4096)=63\).

The evaluation section \ref{evaluation} shows the numerical improvements from enabling TiledMLP with full models.

\hypertarget{ulysses-sp-for-leveraging-aggregate-gpu-memory-across-gpus-for-activations}{%
\subsection{Ulysses Sequence Parallelism for Cross-GPU Activation Memory Sharing}\label{ulysses-sp-for-leveraging-aggregate-gpu-memory-across-gpus-for-activations}}

Now let's have a more detailed look at how UlyssesSPAttentionHF (Ulysses Sequence Parallelism Attention for Hugging Face) works.

\begin{enumerate}
\def\labelenumi{\arabic{enumi}.}
\item Starting with sequence parallelism, the sequence is split across participating GPUs
  and executed in parallel till the attention block is reached.
\item Since self-attention requires an entire sequence, at this boundary we
  switch from sequence parallelism to attention head parallelism
\item When attention block completes we switch back to sequence parallelism
\end{enumerate}

We will use the following diagram (Figure \ref{fig:fig5}) from the
\href{https://www.snowflake.com/en/engineering-blog/ulysses-low-latency-llm-inference/}{\underline{Arctic
Ulysses Inference blog post}} to explain how this works in details:

\begin{figure}[H]
  \centering
  \includegraphics[width=0.9\textwidth]{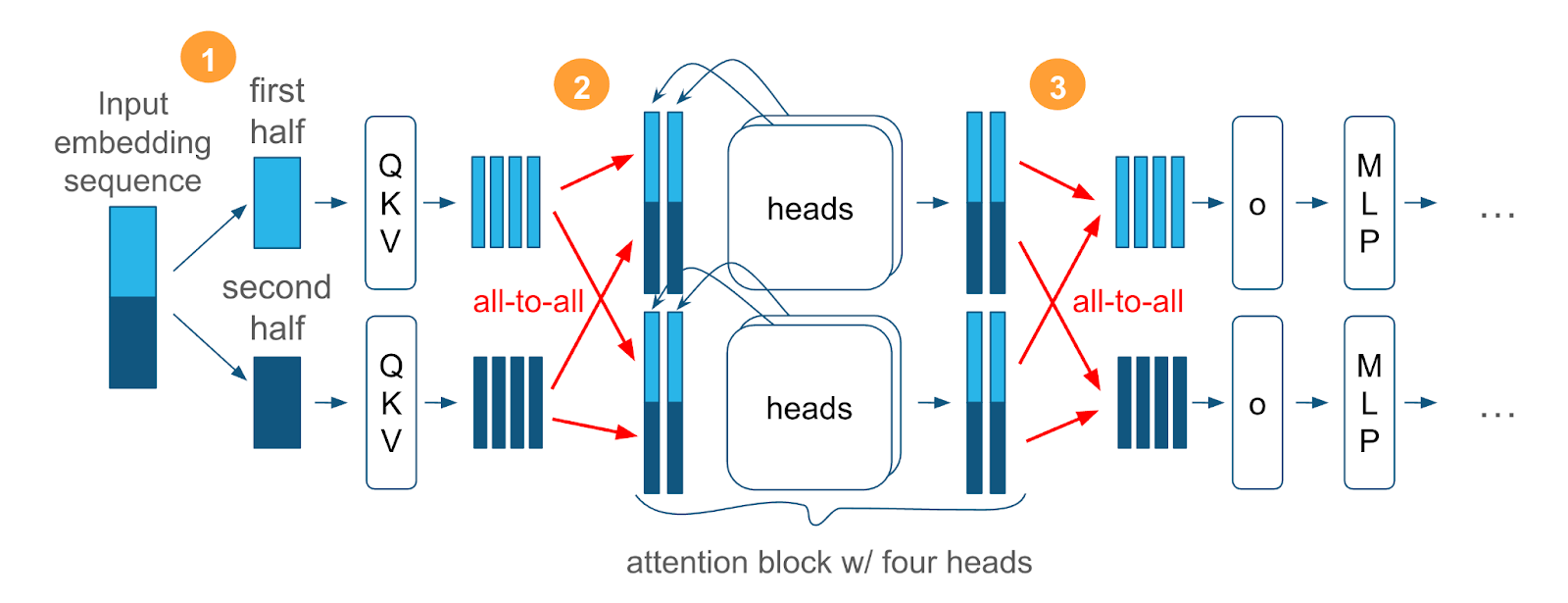}
  \caption{Ulysses Sequence Parallelism diagram with 4 attention heads per attention block model.}
  \label{fig:fig5}
\end{figure}

Since communications of the attention head projections cannot be overlapped
with compute, they have to be really fast. And that's why all-to-all
communication collective is used. Before the all-to-all communication, each rank
has a partial sequence for all attention heads; however, after the
all-to-all communication, each rank has the entire sequence, but only
for a partial subset of attention heads. This allows each rank to
compute the attention for the subset of the attention heads that it owns
in parallel. Then after the attention, Ulysses SP performs another all-to-all communication to switch back to the original
SP layout, where each rank once again has the full embedding (attention
heads) but a shard of a sequence. The computation then proceeds in
parallel until the next layer's attention is reached.

The reason Ulysses SP is attention algorithm-agnostic is because
at the point of calculating the attention it recomposes the full
sequence length and passes it to the desired attention mechanism (e.g.,
FlashAttention2 \cite{dao2023flashattention2fasterattentionbetter} or SDPA). Whereas in other SP approaches (e.g., Ring Attention \cite{liu2023ringattentionblockwisetransformers}) the
attention mechanism itself must be adapted to the specific attention
algorithm being used.

\hypertarget{extending-ulysses-for-modern-attention-mechanism}{%
\subsubsection{Extending Ulysses for Modern Attention Mechanism}\label{extending-ulysses-for-modern-attention-mechanism}}

\begin{figure}[H]
  \centering
  \includegraphics[width=0.8\textwidth]{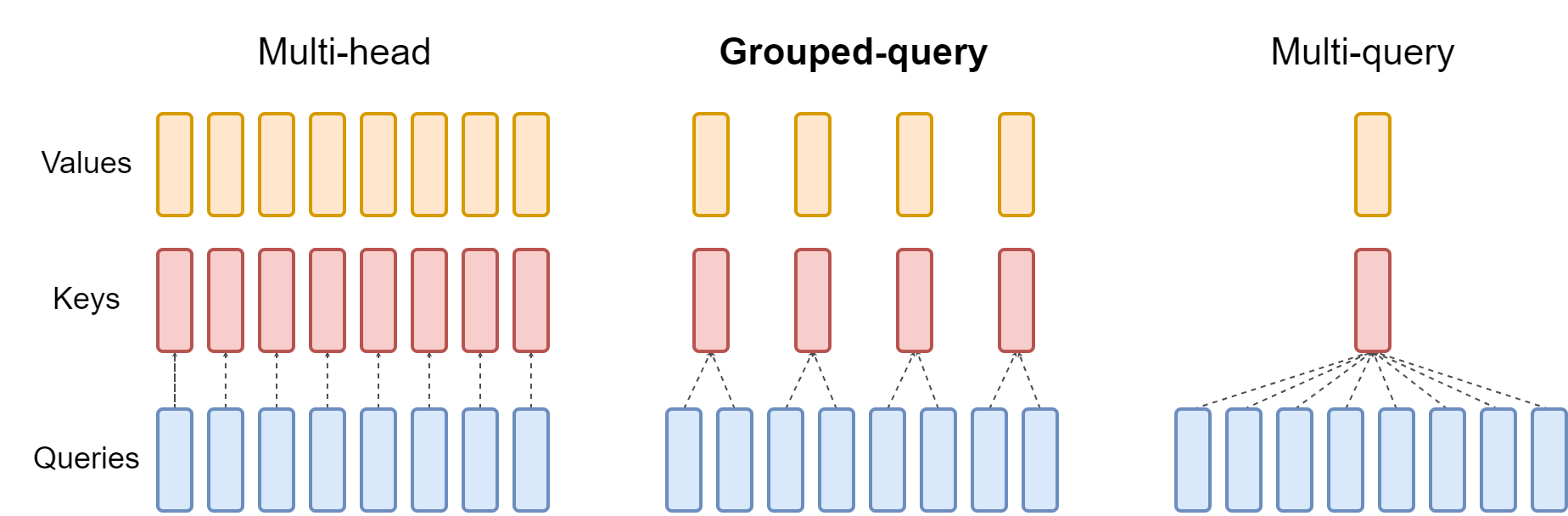}
  \caption{MHA, GQA and MQA types of model attention heads
    (\href{https://arxiv.org/abs/2305.13245}{\underline{source}})
  }
  \label{fig:fig6}
\end{figure}

Figure \ref{fig:fig6} provides a visual representation of MHA, GQA and MQA types of model
attention heads.

The original Ulysses SP implementation in \href{https://github.com/deepspeedai/Megatron-DeepSpeed}{Megatron-DeepSpeed} only supports the MHA type of
attention mechanism. Ulysses SP for HF was extended to support all three of the
aforementioned types:

\begin{enumerate}
\def\labelenumi{\arabic{enumi}.}
\item
  MHA is the simplest since it has the same number of q and kv heads. We
  simply split qkv\_heads into SP shards. Clearly here the only
  limitation is that qkv\_heads is divisible by SP degree.
\item
  GQA type has kv \textless{} q number of heads, since the kv heads get
  reused.

  \begin{enumerate}
  \def\labelenumii{\alph{enumii}.}
  \item
    If kv\_heads is divisible by SP we shard q\_heads into SP degree shards and
    kv\_heads into SP degree shards.
  \item
    If kv\_heads \textless{} SP then we replicate kv\_heads to match SP degree.
  \end{enumerate}
\item
  MQA is where there is 1 kv\_head and many q\_heads. This is the same
  as 2b: we replicate kv\_heads to match SP degree.
\end{enumerate}

Examples:

\begin{itemize}
\item 32 q\_heads, 8 kv\_heads, sp=8 =\textgreater{} each rank will have
  4 q\_heads, 1 kv\_heads
\item 32 q\_heads, 8 kv\_heads, sp=32 =\textgreater{} each rank will
  have 1 q\_heads, 1 kv\_head (kv\_heads will be replicated)
\item 32 q\_heads, 4 kv\_heads, sp=8 =\textgreater{} each rank will have
  4 q\_heads, 1 kv\_heads (kv\_heads will be replicated)
\end{itemize}

The kv\_heads and q\_heads count isn't always divisible by the
desired SP degree. For example, if a model has kv\_heads=3, q\_heads=9
we can do SP=3 or SP=9 and not deploy the node fully to do SP=8*nodes.
\footnote{In the future we plan to come up with solutions that will be able to fit
these nicely into 8x mode in a balanced compute.}

\hypertarget{activation-offloading-and-other-pytorch-optimizations}{%
\subsection{Activation Offload to CPU and Other PyTorch
Optimizations}\label{activation-offloading-and-other-pytorch-optimizations}}

We employed several strategies to reduce runtime memory overhead. Here are the non-invasive ones:

\begin{itemize}
\item A deep memory analysis revealed that PyTorch versions have a
  significant impact on memory usage; due to a known issue with dist.barrier, we
  observed an excess memory usage of over 3 GiB in versions 2.6.0--2.7.0 and
  therefore standardized on version 2.8.0.dev20250507 (aka nightly) for
  our experiments. The just released torch==2.7.1 should be a solid alternative, where this and a few other related problems have been fixed.

\item Additionally, while all\_reduce\_object offers convenience, it
  introduces an additional memory cost of over 3 GiB per GPU, so we opted
  to use all\_reduce instead. 
\item We activated PyTorch activation checkpointing to
  reduce memory usage, accepting a modest increase in compute overhead.
\item Finally, to mitigate memory fragmentation, we enabled
  \href{https://pytorch.org/docs/stable/notes/cuda.html\#optimizing-memory-usage-with-pytorch-cuda-alloc-conf}{\underline{PyTorch's expandable segments allocator}}, which provided massive memory allocation improvements with
  minimal impact on overall performance.
\end{itemize}

Then we introduced a more invasive technique:

The activation checkpointing feature saves an insane amount of memory, but at large
sequence lengths the checkpointed hidden\_states tensors, which are of shape {[bs, seqlen,
hidden\_size]} still consume a lot of GPU memory. For example, at
seqlen=125K/bs=1/hidden\_size=4096/n\_layers=32 it's 30.5GiB across all layers
(\(125\_000\times4096\times2\times32/2^{30}=30.5\)\footnote{Please note that in our experiments we always use batch size 1 when wanting a very long sequence length, because a larger batch size will require a lot more memory.}. We monkey patched
torch.utils.checkpoint.CheckpointFunction to offload the hidden\_states
activation checkpoint tensor to CPU - thus enabling a dramatically
longer sequence length. \footnote{This feature isn't optimized for speed yet - it needs to be switched from direct copy to asynchronous CUDA
streams to allow overlap with forward compute. For
backward it won't help much since it will need to wait till the data is
copied back to GPU for compute to proceed - we are discussing with the
PyTorch team how this feature could be implemented in the core and
enabled with a simple flag in Hugging Face Transformers' from\_pretrained which would make it much simpler for all users.}.

\begin{figure}[H]
  \centering
  \includegraphics[width=1\textwidth]{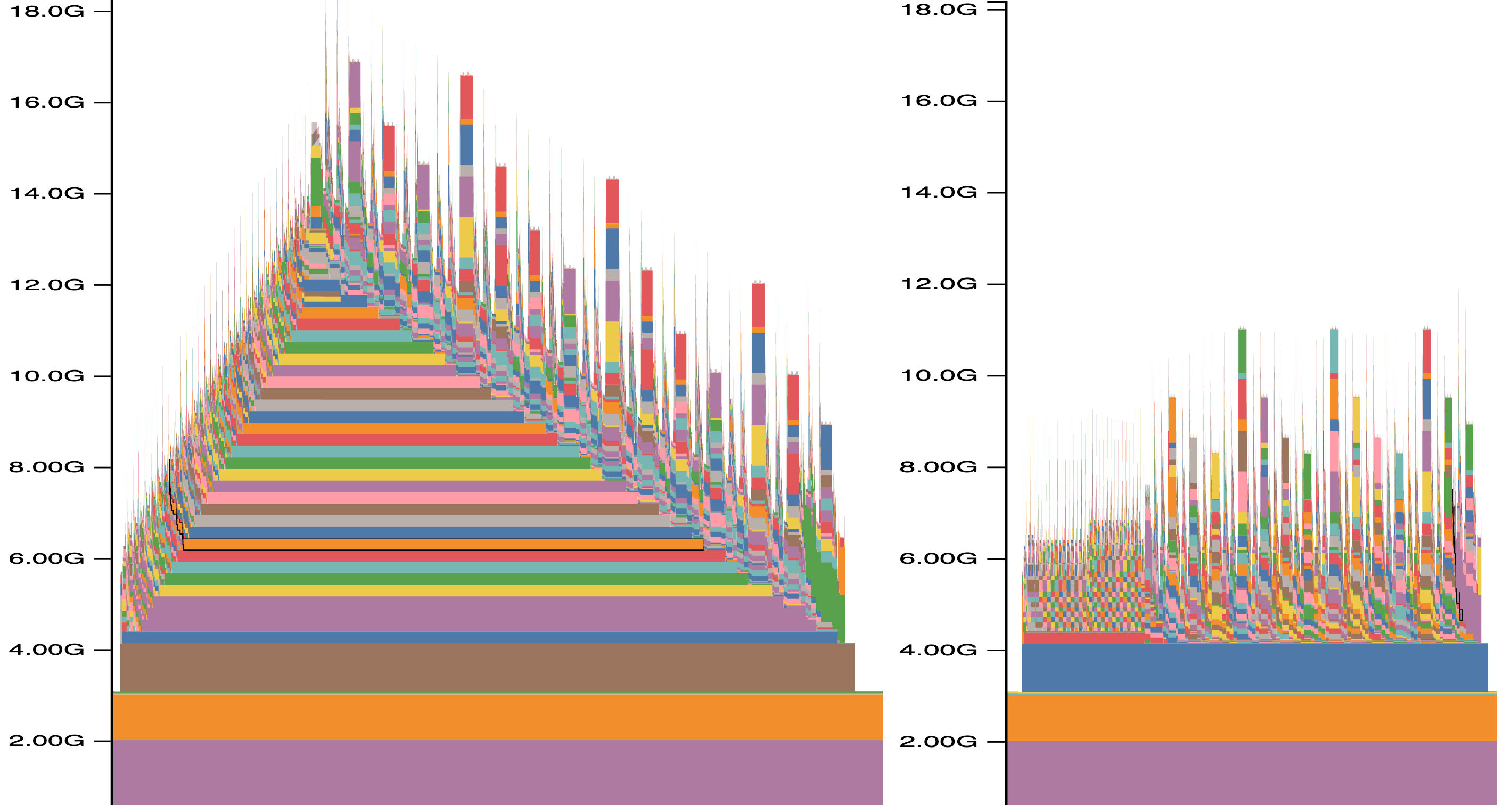}
  \caption{PyTorch memory profiler 1 iteration forward-backward CUDA memory usage: Left: normal setup. Right: with activation checkpoint offloading to CPU enabled}
  \label{fig:fig7}
\end{figure}

Figure \ref{fig:fig7} shows a PyTorch memory profiler visualization of a single forward-backward iteration of Llama-8B with 32k sequence length. On the left you can see the usual pattern of memory usage growing with each layer running its forward calls (left upward slope), followed by memory usage going down during backward calls per layer (right downward slope), when the checkpointed tensors get released once they are no longer needed. On the right, this is the exact same setup, but with activation checkpoint offloading to CPU enabled. It's very clear to see that the "hill" is gone and we are now dealing with a flat structure, which leaves a lot more working GPU memory and allows for a much longer sequence lengths since the peak memory usage is no longer dependent on how many layers the model has.

If you use a large model like Llama-70B you'd need to make sure you
have sufficient CPU memory since hidden\_states will get large and there
are as many copies of this tensor as the number of layers. For example, Llama-70B
at seqlen=3M, bs=1 and 32 GPUs needs 915GiB of CPU memory per node just
for the activation checkpoint offloads
(\(3\_000\_000/32\times8192\times80\times2/2^{30}\times8=915GiB\)) (where hidden\_size=8192,
num\_layers=80). So in this case the node's CPU memory is likely to be the
limiting factor preventing an even bigger sequence length.

This feature can also serve smaller sequence lengths when a very large
batch size is wanted.

\hypertarget{d-attention-mask-is-not-possible-with-long-sequence-lengths}{%
\subsection{4D Attention Mask is not possible with long sequence
lengths}\label{d-attention-mask-is-not-possible-with-long-sequence-lengths}}

When using very long sequence lengths and packed samples deploying \href{https://huggingface.co/blog/poedator/4d-masks}{a 4D causal attention mask} is not
feasible because that tensor is of shape of {[}bs, seqlen, seqlen{]} so
at bs=1 and seqlen=125K it requires a 29GiB tensor
(\(125\_000\times125\_000\times2/2^{30}=29GiB\)) and at 250K it would need 116GiB tensor
on each GPU (\(250\_000\times250\_000\times2/2^{30}=116GiB\)) and it grows quadratically
with sequence length. Therefore, the only way to make the self-attention
attend correctly to sub-samples in the batch without introducing a huge
overhead is to use position\_ids which are of shape {[}bs, seqlen{]}, so
in the case of 125K, it's just 0.2MiB (\(125\_000\times2/2^{20}=0.2MiB\)).

Since currently we can't tell Hugging Face Transformers not to create the causal
mask, other than when using Flash Attention 2, we had to monkey patch
\_update\_causal\_mask so that it won't create this
mask using:

\begin{spverbatim}
           model_without_head = self.model_unwrapped.model
           if hasattr(model_without_head, "_update_causal_mask"):
               model_without_head._update_causal_mask = lambda *args: None
\end{spverbatim}

\hypertarget{implementation-and-hf-transformers-integration-challenges}{%
\section{Implementation and Hugging Face Transformers Integration
Challenges}\label{implementation-and-hf-transformers-integration-challenges}}

\hypertarget{challenges-of-implementing-sequence-parallelism}{%
\subsection{Challenges of Implementing Sequence
Parallelism}\label{challenges-of-implementing-sequence-parallelism}}

We encountered three primary challenges when implementing sequence
parallelism with Hugging Face Transformers models:

\begin{enumerate}
\def\labelenumi{\arabic{enumi}.}
\item
  \begin{quote}
  integrating with existing attention implementations
  \end{quote}
\item
  \begin{quote}
  long-sequence data loading
  \end{quote}
\item
  \begin{quote}
  loss sharding to avoid memory explosion
  \end{quote}
\end{enumerate}

At its core, Ulysses Sequence Parallelism is designed to compose with
existing attention implementations such as SDPA, Flash Attention 2, and
others. While integration is straightforward in frameworks like
Megatron-DeepSpeed, where the integration is done manually in the core,
our approach focuses on extending existing attention modules within the
training framework. This allows support for longer sequence lengths
without requiring changes to the model's code itself.

Long-sequence data loading is particularly challenging, as each training
sample is inherently large. If processed naively, this can lead to
memory exhaustion---exactly what sequence parallelism aims to prevent.
Our solution needed to handle these large sequences efficiently while
remaining compatible with popular dataset providers such as Hugging Face
Datasets~\cite{lhoest-etal-2021-datasets}.

Implementing loss sharding using Sequence Tiling (as introduced in section \ref{sequence-tiling-for-reducing-activation-memory-footprint})
required careful design. The goal was to avoid manual user intervention
and prevent the need for modifications to model implementations which
are outside of our control.

\hypertarget{integration-with-hugging-face-transformers}{%
\subsection{Integration with Hugging Face
Transformers}\label{integration-with-hugging-face-transformers}}

We address these challenges through the following implementation spread out
across Arctic Training~\cite{arctictraining}, DeepSpeed~\cite{deepspeed}, and in some cases small changes to Hugging Face
Transformers~\cite{wolf-etal-2020-transformers} itself.

\begin{enumerate}
\def\labelenumi{\arabic{enumi}.}
\item
  \textbf{Hugging Face Transformers injection} - Ulysses Sequence Parallelism Attention for
  Hugging Face (UlyssesSPAttentionHF) integrates into the modeling code by overriding
  the user-specified \href{https://huggingface.co/docs/transformers/v4.51.3/en/main_classes/model\#transformers.PreTrainedModel.from_pretrained.attn_implementation}{\underline{attn\_implementation}}
  (e.g., sdpa, flash\_attention\_2) with ulysses, and injecting a custom
  wrapper into the Transformers backend via
  transformers.modeling\_utils.ALL\_ATTENTION\_FUNCTIONS. This approach
  allows us to seamlessly wrap the user's intended attention
  implementation with Ulysses SP long-sequence support.
\item
  \textbf{DataLoader} - We introduce a specialized DataLoader adapter,
  UlyssesSPDataLoaderAdapter, which takes any existing DataLoader and
  automatically shards each batch along the sequence dimension. It then
  uses a single rank's batch and processes it collaboratively across all
  ranks---effectively implementing a
  sequence-parallelism-over-data-parallelism protocol. In this setup, we
  iterate over ranks, processing one batch at a time using all ranks in
  parallel. This design preserves the traditional, iterative behavior of
  the original DataLoader, while enabling seamless integration with
  Ulysses Sequence Parallelism.
\item
  \textbf{Non-Attention blocks} - Each rank processes its shard of the
  input sequence independently up to the attention layer, as the
  preceding computations have no inter-token dependencies and can be
  executed in parallel.
\item
  \textbf{Attention block} - Details on how Ulysses handles attention
  can be found in section \ref{ulysses-sp-for-leveraging-aggregate-gpu-memory-across-gpus-for-activations}.
\end{enumerate}

\hypertarget{loss-sharding-in-hf-transformers}{%
\subsection{Loss Sharding in Hugging Face Transformers}\label{loss-sharding-in-hf-transformers}}

In causal language models (e.g., GPT and Llama) cross-entropy loss requires
labels to be shifted one position to the left because of how next-token
prediction works. Cross-entropy loss compares the model's prediction at
its current position to the correct token at the next position.

When computing loss in an unsharded batch we end up with (shift left):

\begin{spverbatim}
input_ids   : [1 2 3 4 5 6 7    8]
labels      : [1 2 3 4 5 6 7    8]
shift_labels: [2 3 4 5 6 7 8 -100]
\end{spverbatim}

-100 is the special label value to be ignored so it gets pushed on the
right.

If we naively shard on the sequence dimension (SP=2 in the following
example), we end up losing the first label of each shard due to Hugging Face Transformers' loss function
shifting each rank's inputs without being aware of our sharding
strategy. This results in our example dropping the token id 5 entirely:

\begin{spverbatim}
input_ids   : [1 2 3    4] [5 6 7    8]
labels      : [1 2 3    4] [5 6 7    8]
shift_labels: [2 3 4 -100] [6 7 8 -100]
\end{spverbatim}

To address this issue,
\href{https://github.com/Hugging Face/transformers/pull/36607}{\underline{we
have modified the causal loss API}} in Hugging Face Transformers to support
user-provided pre-shifted labels. Now we can pre-shift the labels before
sharding on the sequence dimension, and thus end up with the correct
labels passed to the loss function for each shard:

\begin{spverbatim}
input_ids   : [1 2 3 4] [5 6 7    8]
labels      : [1 2 3 4] [5 6 7    8]
shift_labels: [2 3 4 5] [6 7 8 -100]
\end{spverbatim}

\hypertarget{evaluation}{%
\section{\texorpdfstring{Evaluation\\
}{Evaluation }}\label{evaluation}}

\hypertarget{overview}{%
\subsection{Overview}\label{overview}}

We evaluated the longest sequence length we could get for a range of
configurations from 1 GPU to 64 GPUs (8 nodes)\footnote{While ensuring the loss was valid. We have noticed that in some situations when trying to use the last few GiBs of GPU memory the loss would go to NaN.}. Several iterations were completed with each reported sequence length to ensure there are no
memory leaks.

The hypothesis was that, once the minimal number of GPUs required
to hold the model weights is met --- without taking over the whole GPU memory ---
the maximum supported sequence length will scale linearly with the number
of GPUs. And our results confirmed this hypothesis. 

By sharding model weights using ZeRO Stage 3 \cite{rajbhandari2020zeromemoryoptimizationstraining}, each additional GPU reduces the memory load per device, freeing up more memory for activation storage and enabling longer sequence lengths.  In some cases, we even observed a slightly superlinear increase in maximum sequence length.

However, in a few scenarios where the amount of CPU offloading was very large, we run
into a bottleneck of not having enough CPU memory to show the full
potential of these techniques. 

The evaluation results are broken down into 3 sections:

\begin{enumerate}
\item Maximum achieved sequence length - section \ref{maximum-achieved-sequence-length}
\item Feature ablations - section \ref{feature-ablations}
\item Sequence length improvements over baseline - section \ref{sequence-length-improvements-over-baseline}
\end{enumerate}

As the initial goal was to enable long sequence lengths for post-training,
which usually takes just a few days of compute time, we weren't too concerned with the best performance,
but focused primarily on the longest sequence length we could achieve in
reasonable time, offloading things when it was helpful and not doing
that when it was badly impacting the performance. Subsequent work will
focus on improving the performance further.

\hypertarget{methodology}{%
\subsection{Methodology}\label{methodology}}

We will use 1 to 8 node configurations to run the experiments. Each node
is made of 8x H100 80GiB. The nodes are interconnected with EFA v2 (AWS)
that can do \textasciitilde200GBps of all-reduce throughput. The
intra-node connectivity is 450GBps NVLink-4.

Software versions used:

\begin{itemize}
\item
  torch==2.8.0.dev20250507 (a.k.a. torch-nightly)\footnote{we found that torch<2.7.1 versions
  either had memory leaks or weren't as memory-efficient.} torch>=2.7.1 should be as good\footnote{torch==2.7.1 version had been released when this paper was about to be completed.})
\item
  flash\_attn==2.7.4 (flash\_attn>=2.6.4 delivers a very similar performance)
\item
  transformers==4.51.3 (transformers>=4.51.3 should be fine)
\item
  deepspeed==0.17.0 (deepspeed>=0.17.0 should be fine)
\end{itemize}

The following optimizations were enabled during all, but ablation experiments:

\begin{itemize}
\item \href{https://www.deepspeed.ai/tutorials/zero/}{\underline{DeepSpeed ZeRO
  Stage 3}}
\item \href{https://www.deepspeed.ai/2021/03/07/zero3-offload.html}{\underline{DeepSpeed
  optimizer states offload to CPU}}
  \item \href{https://github.com/stas00/ml-engineering/blob/master/training/performance/README.md#gradient-checkpointing}{\underline{Gradient/Activation checkpointing}}
\item Fused tiled logits+loss computation via \href{https://github.com/linkedin/Liger-Kernel}{\underline{Liger-Kernel}}  \footnote{all of Liger-Kernel's Llama features were enabled, except swiglu when TiledMLP was used, because Liger-Kernel monkey patches LlamaMLP as well when swiglu is enabled, leading to a conflict with TiledMLP.}
\item \href{https://pytorch.org/docs/stable/notes/cuda.html\#optimizing-memory-usage-with-pytorch-cuda-alloc-conf}{\underline{PYTORCH\_CUDA\_ALLOC\_CONF=expandable\_segments:True}}
  environment variable
\item Sequence parallel communications were done in bf16 (or could reduce the communication
  buffer size instead)
\item Tiled MLP computation
\item Activation checkpoint hidden\_states tensor offload to CPU
\end{itemize}

Additionally, when we used a single GPU, we also enabled model weights offload to CPU to prevent GPU OOM, which otherwise would occur even with a tiny sequence length.

\hypertarget{maximum-achieved-sequence-length}{%
\subsection{Maximum Achieved Sequence Length}\label{maximum-achieved-sequence-length}}

We measured the maximum achievable sequence length with three popular representative models by zeroing in on the maximum length that would not provide out of memory, loss=NaN and other errors.

\subsubsection{meta-llama/Llama-3.1-8B-Instruct}

\href{https://huggingface.co/meta-llama/Llama-3.1-8B-Instruct}{\underline{meta-llama/Llama-3.1-8B-Instruct}} has 32 q\_heads and 8 kv\_heads and thus can be trained on 1 to 32 GPUs. 

\begin{figure}[H]
  \centering
  \includegraphics[width=0.8\textwidth]{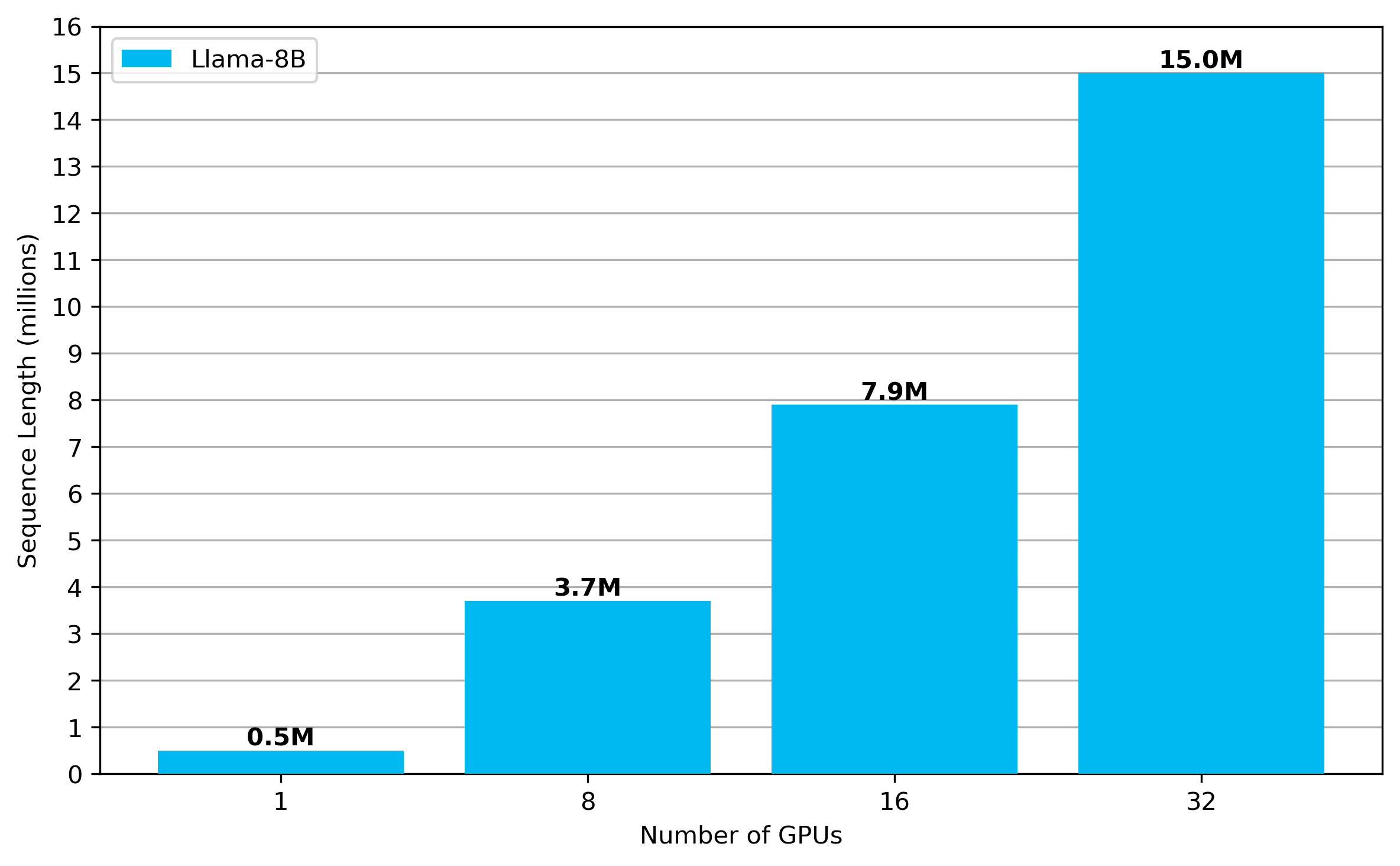}
  \caption{Maximum achieved sequence length with meta-llama/Llama-3.1-8B-Instruct}
  \label{fig:fig8}
\end{figure}

Figure \ref{fig:fig8} shows the measured outcomes.

\subsubsection{meta-llama/Llama-3.1-70B-Instruct}

\href{https://huggingface.co/meta-llama/Llama-3.1-70B-Instruct}{\underline{meta-llama/Llama-3.1-70B-Instruct}} has 64 q\_heads and 8 kv\_heads and thus can be trained on 16 to 64 GPUs. At least 8 GPUs are needed to just fit the sharded model and gradients, while offloading optimizer states to CPU. 

\begin{figure}[H]
  \centering
  \includegraphics[width=0.8\textwidth]{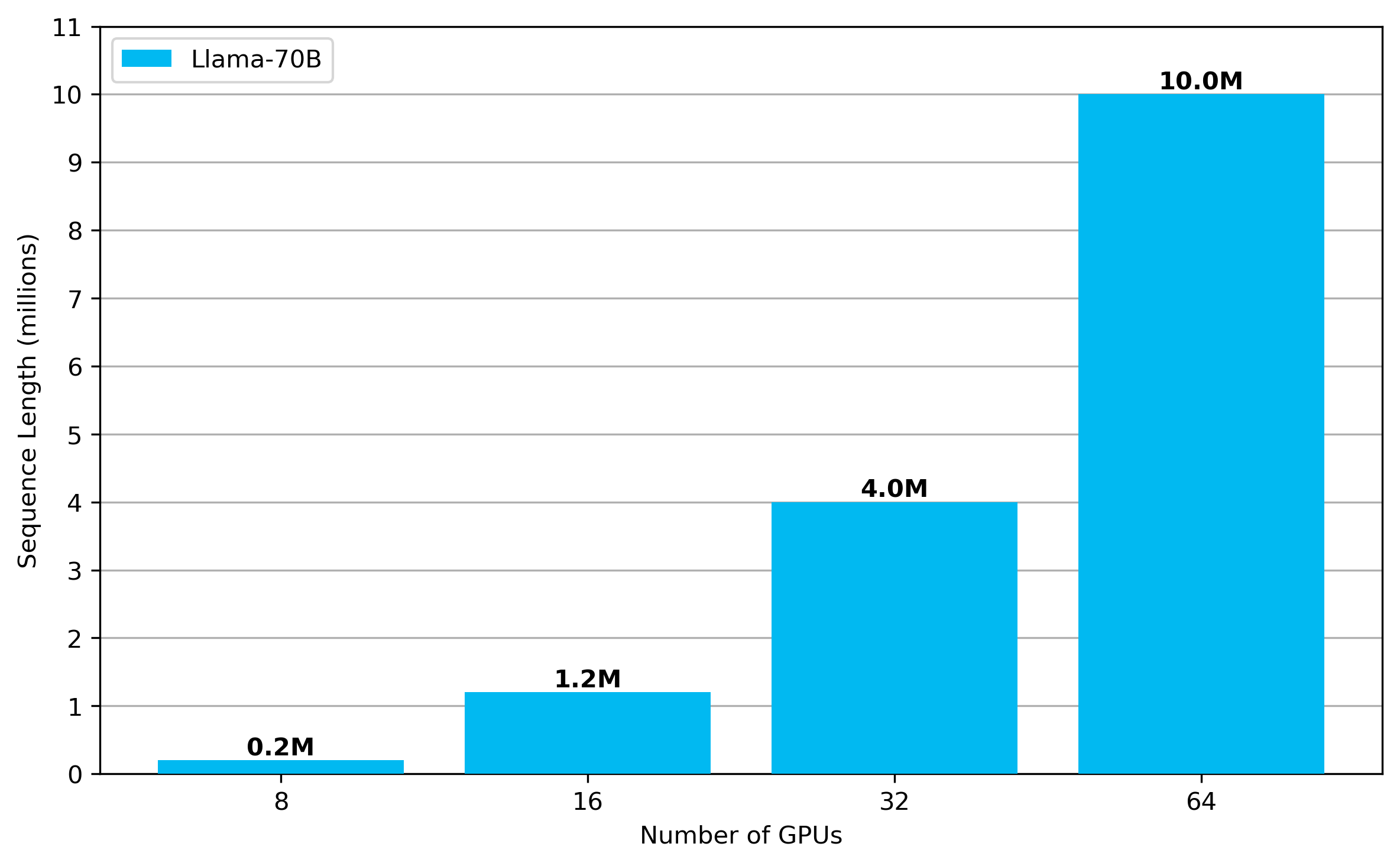}
  \caption{Maximum achieved sequence length with meta-llama/Llama-3.1-70B-Instruct}
  \label{fig:fig9}
\end{figure}

Figure \ref{fig:fig9} shows the measured outcomes.

As you can see from the notes, we couldn't unleash the full sequence length potential for this model because activation checkpoint offload to CPU memory requirements for this model are so big:

\begin{itemize}
\item 4 nodes: \(1\_000\_000/32\times8192\times80\times2/2^{30}\times8=305GiB\) per 1M seqlen
\item 8 nodes: \(1\_000\_000/64\times8192\times80\times2/2^{30}\times8=152GiB\) per 1M seqlen
\end{itemize}

And we had only 1.9TB of cpu memory available per node. We know that we
"left more sequence length on the table", because the GPU memory was only
about ¾ full.

\subsubsection{Qwen/Qwen3-32B}

\href{https://huggingface.co/Qwen/Qwen3-32B}{\underline{Qwen/Qwen3-32B}} has 64 q\_heads and 8 kv\_heads and thus can be trained on 1 to 64 GPUs. 

\begin{figure}[H]
  \centering
  \includegraphics[width=0.8\textwidth]{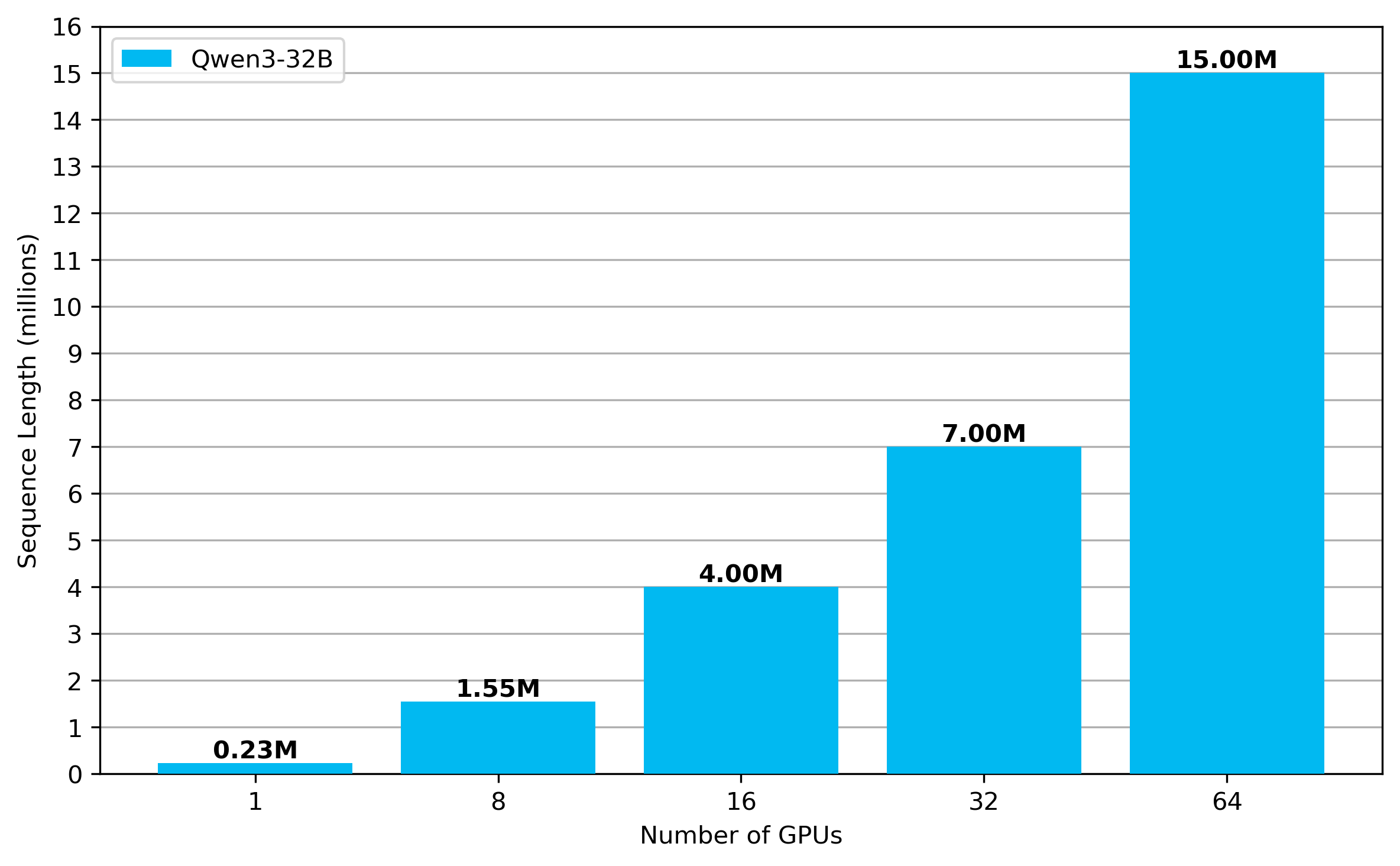}
  \caption{Maximum achieved sequence length with Qwen/Qwen3-32B}
  \label{fig:fig10}
\end{figure}

Figure \ref{fig:fig10} shows the measured outcomes.

For a single GPU an additional weights offload to cpu was required.

Same as with Llama-70B for a few configurations we discovered that 1.9TiB
of CPU memory weren't enough to get an even longer sequence length. For
this model activation checkpoint offload to CPU memory requirements were:

\begin{itemize}
\item  4 nodes: \(1\_000\_000/32\times5120\times64\times2/2^{30}times8=152GiB\) per 1M seqlen
\item  8 nodes: \(1\_000\_000/64\times5120\times64\times2/2^{30}\times8=76GiB\) per 1M seqlen
\end{itemize}

\subsubsection{Summary}

As can be seen from the evaluation numbers for three different models, the possible sequence length
growth is roughly linear, that is doubling the number of nodes, doubles
the possible sequence length. In fact it's a bit better than linear
because of DeepSpeed ZeRO Stage 3, which partitions the model constituents into shards
across all GPUs, so the more nodes are used the smaller the shards owned
by each GPU, and as a result the more GPU memory is available for activations.

\hypertarget{feature-ablations}{%
\subsection{\texorpdfstring{Feature Ablations
}{Feature Ablations }}\label{feature-ablations}}

For the feature ablation experiments we use a single 8x H100 node.

\textbf{Baseline:}

\begin{enumerate}
\item \href{https://www.deepspeed.ai/tutorials/zero/}{\underline{DeepSpeed ZeRO Stage 3}}
\item \href{https://github.com/stas00/ml-engineering/blob/master/training/performance/README.md#gradient-checkpointing}{\underline{Gradient/Activation checkpointing enabled}}
\item \href{https://www.deepspeed.ai/2021/03/07/zero3-offload.html}{\underline{DeepSpeed Optim states offload to CPU}}
\item \href{https://docs.pytorch.org/docs/stable/notes/cuda.html#optimizing-memory-usage-with-pytorch-cuda-alloc-conf
}{\underline{PYTORCH\_CUDA\_ALLOC\_CONF=expandable\_segments:True}}
\item Flash Attention 2 \cite{dao2023flashattention2fasterattentionbetter}
\end{enumerate}

We next perform feature ablations on each of the following features and show the outcome in Table \ref{tab:tab1}:

\begin{enumerate}
\item Fused tiled logits \& loss compute with Liger-Kernel
\item Ulysses Sequence Parallelism for Hugging Face
\item Tiled MLP
\item Activation checkpoint offload to CPU
\end{enumerate}

\setlength{\arrayrulewidth}{0.3mm}
\renewcommand{\arraystretch}{1.5}

\begin{table}[h!]
\centering
\caption{Feature ablations results.}
\begin{tabular}{|C{1cm} C{1cm} C{1.2cm} C{1cm} C{1.6cm}|R{1.2cm}|R{1.3cm}|R{1.3cm}|} 
 \hline
 Base line & Tiled Logits \& Loss & Ulysses SP for HF & Tiled MLP & Activation checkpoint offload to CPU & Sequence length (bs=1) & Iteration time h:mm:ss & TFLOPS \\
 \hline
 \checkmark &            &            &            &            & 32K  & 0:00:17 & 231.6 \\
 \hline
 \checkmark & \checkmark &            &            &            & 160K & 0:02:03 & 514.4 \\
 \hline
 \checkmark & \checkmark & \checkmark &            &            & 1.1M & 0:09:24 & 576.1 \\
 \hline
 \checkmark & \checkmark & \checkmark & \checkmark &            & 1.2M & 0:11:43 & 548.7 \\
 \hline
 \checkmark & \checkmark & \checkmark &            & \checkmark & 2.4M & 0:43:30 & 585.8 \\
 \hline
 \checkmark & \checkmark & \checkmark & \checkmark & \checkmark & 3.7M & 1:47:35 & 590.6 \\
 \hline
\end{tabular}
\label{tab:tab1}
\label{table:1}
\end{table}

\begin{figure}[H]
  \centering
  \includegraphics[width=0.9\textwidth]{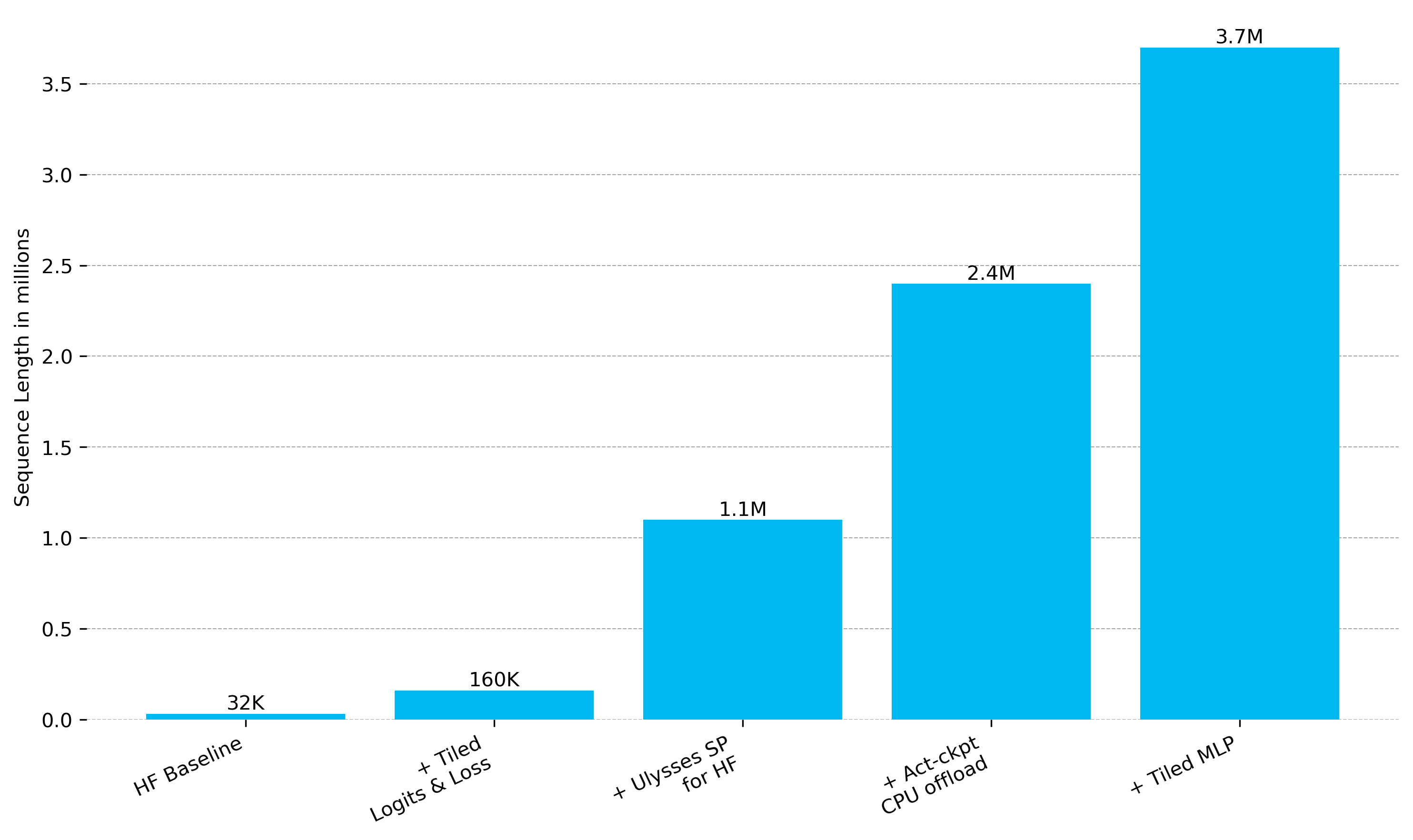}
  \caption{Feature ablation results visualized}
  \label{fig:ablation}
\end{figure}

Table \ref{tab:tab1} and the corresponding Figure \ref{fig:ablation} show that tiling computation features like tiled logits \& loss and tiled MLP don't contribute much at a low
sequence length, but once activation checkpoint offload to CPU enabled a much
larger sequence length, then for example tiled MLP was able to increase the sequence
length by an additional 58\%. Activation checkpoint offload to CPU and tiled MLP together enabled a
sequence length that's \textasciitilde3.5 times larger than the baseline
+ fused tiled logits \& loss (Liger Kernel) + Ulysses SP for HF. Once we hit sequence lengths larger
than 5M tiled MLP starts to massively contribute to allowing a much
larger sequence length. hidden\_states have a {[}bs, seqlen,
hidden\_size{]} shape and while hidden\_size remains the same, seqlen
becomes very big, leading to many GiBs-large hidden\_states tensor for each
layer.

Since the attention computation is quadratic with regards to sequence
length, it's easy to see how the iteration time dramatically slows down
as sequence length grows.

Additional notes:

\begin{itemize}
\item If Liger-Kernel doesn't support the architecture of your need, we have
  implemented \href{https://github.com/deepspeedai/DeepSpeed/blob/097f0637d5b002f8d3e1a647f8622ef0359438a4/deepspeed/runtime/sequence_parallel/ulysses_sp.py#L552}{Sequence Tiled Compute} that can perform a tiled cross-entropy loss that should work with any
  causal model. It saves approximately the same amount of memory but
  it's slower than Liger-Kernel as it's written in pure PyTorch.  It can be found \href{https://github.com/snowflakedb/ArcticTraining/blob/229764ef817db231048f1c5e822726eac2cd12c5/arctic_training/trainer/sft_trainer.py#L108-L151}{here}.
\item When Tiled MLP is enabled Liger-Kernel's swiglu override is turned off
  since the 2 compete with each other over Hugging Face Transformers' modeling MLP class override\footnote{but if one wants Liger-Kernel's efficient
  swiglu it should be possible to combine the two together with some
  additional work.}.
\item The TFLOPS and Iteration time were measured while using non-packed
  samples and using the standard Megatron-LM flos\footnote{to prevent ambiguity we use the abbreviation flos instead of flops, which was coined during BLOOM-176B work\cite{workshop2023bloom176bparameteropenaccessmultilingual}, as flops could mean floating point operations per second and also just floating point operations - but here we refer to the latter.} estimation formulae
  taking into account repeated forwards. At such a long sequence length
  attention computation renders MLP compute negligible. We observed
  packed samples with FlashAttention2 setups report much lower TFLOPS.
\item Since all 8 GPUs compute a single batch, and the micro-batch size is
  1, the effective global batch size is 1 as well.
\end{itemize}

\subsection{Sequence Length Improvements over Baseline}\label{sequence-length-improvements-over-baseline}

\begin{figure}[H]
  \centering
  \includegraphics[width=0.8\textwidth]{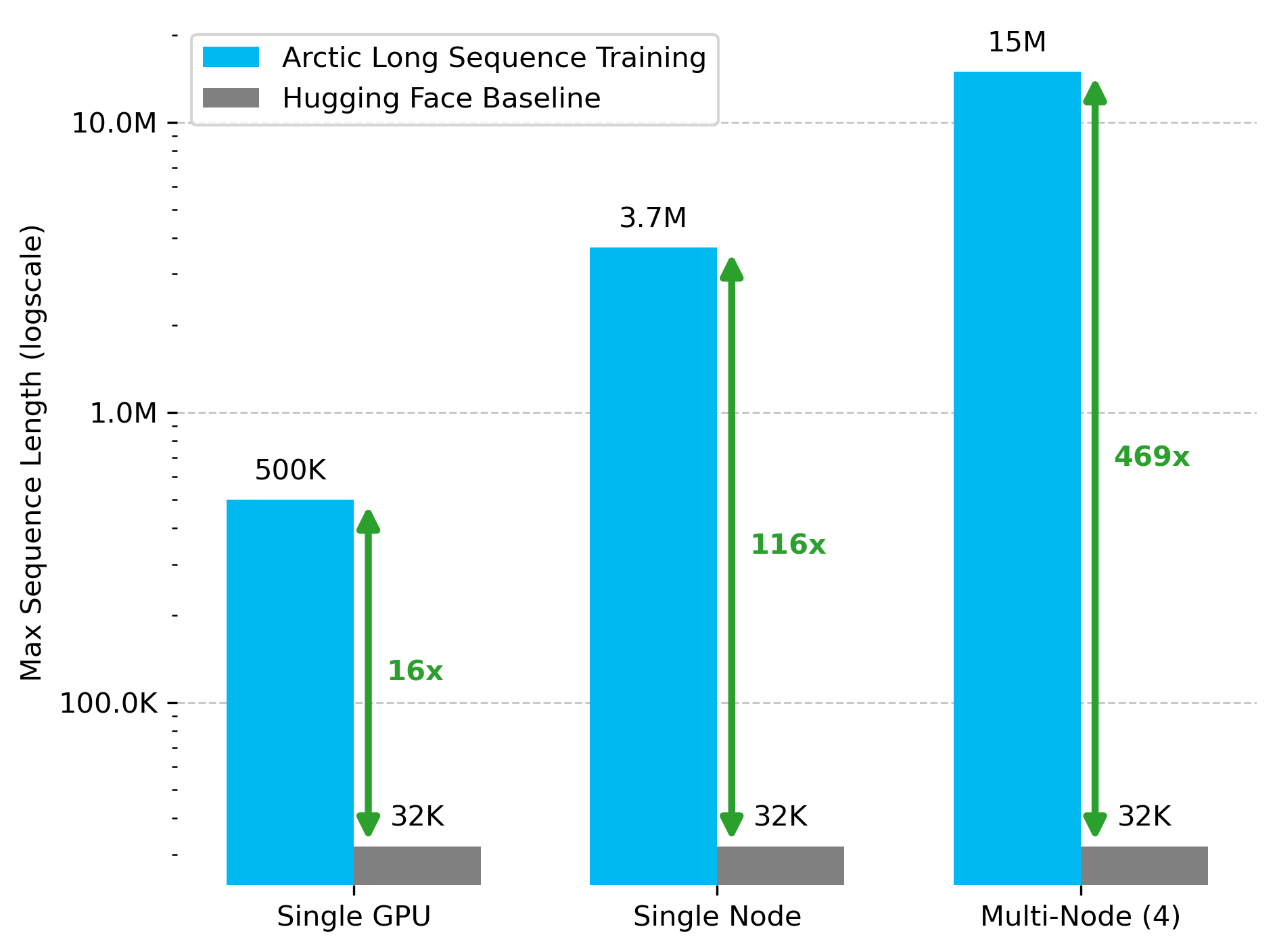}
  \caption{The impact of enabling ALST for LLama-8B on 1, 8 and 32 GPUs}
  \label{fig:fig11}
\end{figure}

After activating ALST with LLama-8B the sequence length improvements were as following:

\begin{itemize}

\item 1 GPU: from 32K\footnote{the same baseline as in all evaluations.\label{baseline_footnote}}\footnote{this setup’s baseline additionally includes model weights offload to CPU, otherwise the setup couldn't fit on a single GPU without OOM.} to 500K, a 16 times improvement, as demonstrated by Table \ref{tab:tab2} and Figure \ref{fig:fig11}.

\item 8 GPUs: from 32K\footref{baseline_footnote} to 3.7M, a 116 times improvement, as demonstrated by Table \ref{tab:tab3} and Figure \ref{fig:fig11}.

\item 32 GPUs: from 32K\footref{baseline_footnote} to 15M, a 469 times improvement, as demonstrated by Table \ref{tab:tab4} and Figure \ref{fig:fig11}.

\end{itemize}

\begin{table}[h!]
\centering
\caption{Sequence length improvement for Llama-8B on a single H100 GPU}
\begin{tabular}{|C{1.3cm} C{1cm}|R{2cm}|R{2cm}|R{1.3cm}|} 
 \hline
 Baseline & ALST & Sequence length (bs=1) & Iteration time h:mm:ss & TFLOPS \\
 \hline
 \checkmark &            &  32K  & 0:00:26 & 189.4 \\
 \hline
 \checkmark & \checkmark &  500K & 0:16:50 & 548.1 \\
 \hline
\end{tabular}
\label{tab:tab2}
\label{table:2}
\end{table}

\begin{table}[h!]
\centering
\caption{Sequence length improvement for Llama-8B on 8 H100 GPUs (1 node)}
\begin{tabular}{|C{1.3cm} C{1cm}|R{2cm}|R{2cm}|R{1.3cm}|} 
 \hline
 Baseline & ALST & Sequence length (bs=1) & Iteration time h:mm:ss & TFLOPS \\
 \hline
 \checkmark &            &  32K & 0:00:17 & 231.6 \\
 \hline
 \checkmark & \checkmark & 3.7M & 1:47:35 & 590.6 \\
 \hline
\end{tabular}
\label{tab:tab3}
\label{table:3}
\end{table}

\begin{table}[h!]
\centering
\caption{Sequence length improvement for Llama-8B on 32 H100 GPUs (4 nodes)}
\begin{tabular}{|C{1.3cm} C{1cm}|R{2cm}|R{2cm}|R{1.3cm}|} 
 \hline
 Baseline & ALST & Sequence length (bs=1) & Iteration time h:mm:ss & TFLOPS \\
 \hline
 \checkmark &            &  32K & 0:00:12 & 393.6 \\
 \hline
 \checkmark & \checkmark &  15M & 7:25:09 & 590.6 \\
 \hline
\end{tabular}
\label{tab:tab4}
\label{table:4}
\end{table}

Please note that sequence length is on the log scale because otherwise the baseline wasn't showing in the 32 GPU plot.

\subsection{Training Correctness}

We used Llama-8B to validate that ALST matches the baseline on training quality. We compared a 32k sequence length on a single 8x H100 GPUs node.

In order to ensure equal conditions, in the case of ALST, since we use 8 GPUs to process a single sample there, we enabled gradient accumulation steps of 8. That way each iteration in both setups has seen the exact same data.

\begin{figure}[H]
  \centering
  \includegraphics[width=0.9\textwidth]{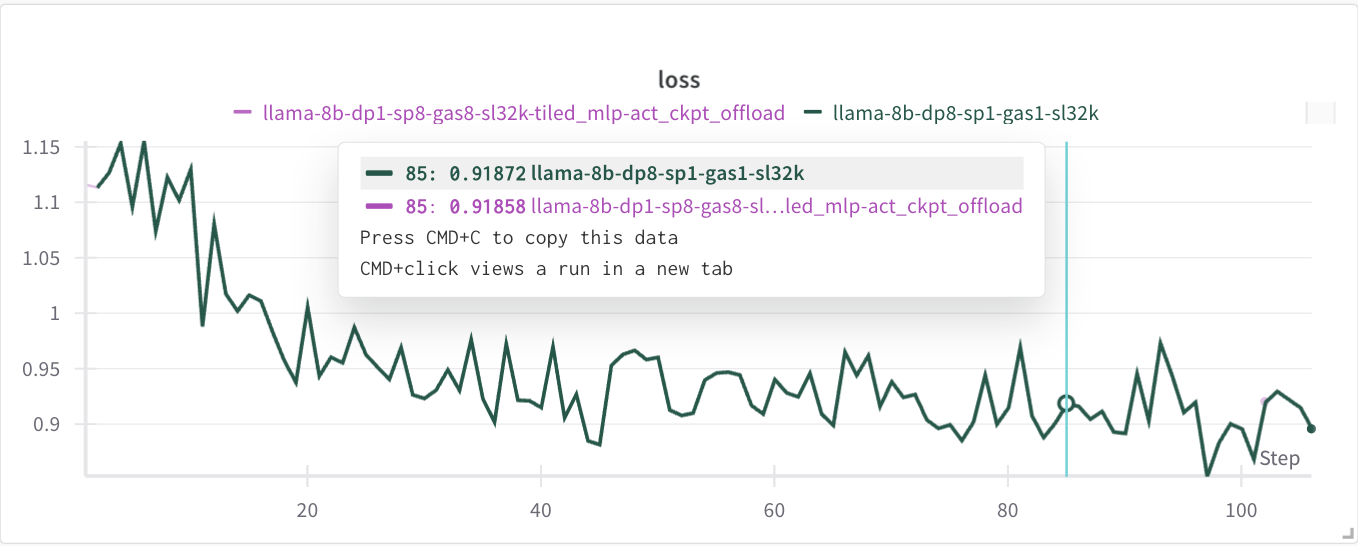}
  \caption{Training loss comparison with and without ALST}
  \label{fig:correctness}
\end{figure}

As can be observed from Figure \ref{fig:correctness} we have an almost exact match for the loss with ALST \footnote{The two plots overlap almost exactly and the tiny differences can only be seen from checking the exact loss values for each iteration in the floating box with loss values per plot.}. Thus we know ALST provides the same training quality as the baseline.

\hypertarget{trying-it-out}{%
\section{\texorpdfstring{Trying it out\\
}{Trying it out }}\label{trying-it-out}}

The
\href{https://github.com/snowflakedb/ArcticTraining}{\underline{ArcticTraining}}
framework has fully working post-training recipes using ALST for a variety of models and quantities of GPUs. You can just drop in your dataset definition and run the post-training.

Go to
\href{https://github.com/snowflakedb/ArcticTraining/blob/main/projects/sequence-parallelism/README.md}{\underline{https://github.com/snowflakedb/ArcticTraining/blob/main/projects/sequence-parallelism/README.md}}
and follow the instructions there to reproduce any of the evaluations presented in this paper or adapt the existing recipes to your long sequence length finetuning needs.

\hypertarget{additional-notes-for-users}{%
\section{Additional Notes for Users}\label{additional-notes-for-users}}

\hypertarget{limitations}{%
\subsection{Limitations}\label{limitations}}

\begin{itemize}
\item
  Currently the maximum degree of sequence parallelism (SP) is limited by the number of
  q\_heads. For example, meta-llama/Llama-3.1-70B-Instruct has 64
  q\_heads, so SP=64 is the maximum possible with that model. We plan to remove that
  limit in future work. Meanwhile you can still scale beyond the SP
  limit imposed by head count, while using a higher DP degree. For
  example, you can easily train on 1024 GPUs, there will be 16 SP
  replicas of SP=64 each.
\item
  As discussed earlier q\_heads need to be divisible by SP degree. For example, if the model has 9 q\_heads, you'd need SP to be 1, 3 or 9. We plan to overcome this limitation in the future.
\end{itemize}

\hypertarget{important-training-notes}{%
\subsection{Important Training Notes}\label{important-training-notes}}

While this technology enables you to train on very long sequences, you
need to be aware that if you pack many short sequences into a long one
it won't learn to infer on long sequences. You need to use a
dataset with samples whose sequence length matches your target goal.
If you train on packed samples, it'd be akin to having a large
batch size of short sequence length samples.

Because the dense attention mechanism has a quadratic O(s\^{}2)
relationship with the sequence length - the longer the individual
sample, the slower the attention calculation will be. As of this
paper's writing the incarnation of Ulysses SP for HF supports both SDPA
and Flash Attention 2 (FA2) as they are integrated into Hugging Face
Transformers. FA2 is very efficient at calculating the attention of
individual samples packed into a long sequence by using position ids,
whereas SDPA in Hugging Face Transformers as of this writing ignores position ids and
ends up attending to the whole packed sequence which is both much
slower and isn't correct. Though as explained earlier, to post-train your model for long
sequence lengths you have to use actual long sequence length samples and not
packed short samples, in which case both SDPA and FA2 will work
correctly.

\section{Future Work}

While large matrix multiplications dominate training with very long sequence lengths, making other operations quite insignificant performance-wise, additional work can be done to further improve the performance of various components where they don't overlap with compute. 

While the initial implementation has been integrated into \href{https://github.com/snowflakedb/ArcticTraining}{\underline{Arctic Training}} - we next would like to integrate it into Hugging Face Accelerate and Trainer and various other frameworks to make it easy for any user to access this technology. The integration document can be found \href{https://www.deepspeed.ai/tutorials/ulysses-alst-sequence-pallellism/}{here}.

\section*{Acknowledgments}

Besides the paper's authors the following folks have contributed to this work and we would like to thank them. The Hugging Face team: Cyril Vallez, Yih-Dar Shieh and Arthur Zucker. The PyTorch team: Jeffrey Wan, Mark Saroufim, Will Constable, Natalia Gimelshein, Ke Wen and alband. This paper's reviewers: Ari Rean and Anupam Datta. Also, we would like to acknowledge the original Ulysses for Megatron-DeepSpeed team: Sam Ade Jacobs, Masahiro Tanaka, Chengming Zhang, Minjia Zhang, Shuaiwen Leon Song, Samyam Rajbhandari and Yuxiong He.

%
\nocite{*}
\bibliography{references}
\bibliographystyle{IEEEtran}
\end{document}